\documentclass[11pt,a4paper]{article}
\usepackage[hyperref]{acl2020}
\usepackage{times}
\usepackage{latexsym}

\usepackage{hyperref}
\usepackage{amsmath}
\usepackage{booktabs}
\usepackage{arydshln}
\usepackage{graphicx}
\usepackage{arabtex}
\usepackage{utf8}

\DeclareMathOperator*{\argmax}{arg\,max}

\usepackage{microtype}

\aclfinalcopy
\def\aclpaperid{2} %  Enter the acl Paper ID here

\title{Stance Prediction and Claim Verification: An {A}rabic Perspective}

\author{Jude Khouja \\
  Latynt\\
  \texttt{jude@latynt.com} \\}

\date{5/15/2020}

\begin{document}
\maketitle

% Because of package arabex, we need to add /end{list} 
\begin{abstract}
This work explores the application of textual entailment in news claim verification and stance prediction using a new corpus in Arabic. The publicly available corpus comes in two perspectives: a version consisting of 4,547 true and false claims and a version consisting of 3,786 pairs (claim, evidence). We describe the methodology for creating the corpus and the annotation process. Using the introduced corpus, we also develop two machine learning baselines for two proposed tasks: claim verification and stance prediction.  Our best model utilizes pretraining (BERT) and achieves 76.7 \emph{F1} on the stance prediction task and 64.3 \emph{F1} on the claim verification task.  Our preliminary experiments shed some light on the limits of automatic claim verification that relies on claims text only. Results hint that while the linguistic features and world knowledge learned during pretraining are useful for stance prediction, such learned representations from pretraining are insufficient for verifying claims without access to context or evidence.
\end{list}\end{abstract} 

\section{Introduction}
Although fake news is not an emerging phenomenon and has been documented throughout history, the prevalence and wide spread of misinformation over the internet has captured significant proportion of public attention in recent years.  This is in part linked to the low barrier for content generation through the advent of the internet and social media \citep{ Allcott2017-cz} and the fact that false news spread faster than true news \citep{Vosoughi2018-oi} rendering it increasingly dangerous to public discourse. The widespread exposure in the U.S. for example has been reported by researchers who found that the average  American  encountered  between  one  and  three  stories  from  known  publishers  of  fake  news  during  the  month  before  the 2016 election \citep{Allcott2017-cz}.

Since manual fact-checking by human experts does not scale well with the amount of information shared on the web, there is a growing body of work in recent years aimed at developing automatic tools to target fake news, misinformation and credibility of content on social media in general \citep{Rubin_undated-kb, el-ballouli-etal-2017-cat, Baly2018-ij, Baly2018-ub, Yaqing_Wang_SUNY_Buffalo_Buffalo_NY_USA_undated-bt, Saleh2019-sv, Zhang2019-od}.  Several datasets were developed to further aid research on this topic\footnote{FNC: http://www.fakenewschallenge.org/} \citep{Darwish2017-ca, Wang2017-iy, Baly2018-ub, Thorne2018-iw}. We refer readers to \citep{Thorne2018-uf, Pierri2019-bc} for a more comprehensive overview of recent research on fake news, propaganda and misinformation.

Despite the increased attention, most of the work has been focusing on the English language. Tools, resources and datasets available in Arabic are limited \citep{Darwish2017-ca, Baly2018-ub, Elsayed2019-ma}. As such, this work contributes to recent efforts targeting Arabic by introducing a new publicly available corpus in Arabic that is suitable to study claim verification and semantic entailment \citep{Katz1972-iu}.

\section{Related Work}
\label{sec:related}
In recent years, there has been rapid progress in developing systems and tools for automatic fact checking and claim verification.  Various approaches were developed which relied on a diverse set of methods and information to verify claims.  Most relevant to this work are approaches that used content such as textual information in the title and/or body of the claims to predict their veracity. Among this direction of research those that considered a machine learning approach \citep{Potthast2017-ra, Yaqing_Wang_SUNY_Buffalo_Buffalo_NY_USA_undated-bt, Alzanin2019-qo} including deep learning techniques \citep{Hanselowski2017-zf, Baly2018-ub, Popat2018-pw, Chawla2019-hp, Helwe2019-yi, Lv2019-sp}.

\paragraph{Datasets:}There are limited but growing datasets related to claim verification \citep{Al_Zaatari2016-pn, Darwish2017-ca, Wang2017-iy,   Baly2018-ub, Thorne2018-iw,  Alkhair2019-cz, Alzanin2019-qo, Elsayed2019-ma}. However, datasets focusing on Arabic remain scarce \citep{Darwish2017-ca, Baly2018-ub, Elsayed2019-ma}. Recently, work on the application of textual entailment for claim verfication has been explored and new datasets combining stance prediction and claim verfication were introduced \citep{Baly2018-ub, Thorne2018-iw}. 

\begin{table}
    \centering
    \scalebox{0.74}{%
    
     \begin{tabular}{l}
     \toprule
        Given a news title, write two news titles that: \\
        \\
        A- Paraphrase the original title:\\
        Has same meaning but is worded differently by\\
        rephrasing and changing Syntax, using verb\\
        synonyms, using different words to describe the\\
        same information such as locations, counts and dates.
        \\
        \\
        B- Contradict the original title:\\
        Looks similar to the original title but has\\
        contradicting meaning (both cannot be true in the\\
        same context) by reversing meaning without\\
        negating main verb, using antonym of main verb\\
        with rephrasing, changing key information using world\\
        knowledge such as locations, counts and dates.
    \\\hline
    \end{tabular}%
    }
     \caption{\small{Guidelines for rewriting news titles.}}
     \label{tab:guides}
\end{table}

\paragraph{}
This work is most in line with that direction. We developed a new corpus in Arabic that can be used jointly for claim verification and textual entailment recognition. However, our new corpus differs from the aforementioned datasets in that it is at the sentence level, hence, we are disentangling the tasks of claim verification and textual entailment from the task of evidence extraction (Information Retrieval) and focusing on the former.  We also start from real news titles and generate true/false claims from them. Our aim is to mitigate one type of bias that results from starting with fake news collected in the wild: bias in the distribution of topics among the true/false claims.  While some forms of biases about the world are useful in determining the veracity of a claim, some can be problematic. We can imagine a dataset that contains more positive\footnote{Positive here refers to sentiment} news in the "fake" class than in the "true" class.  A system trained on such data could predict the class "fake" with higher confidence for any claim that has a positive tone compared to one that has a negative or neutral tone.  Such surface level biases in topics and linguistic styles could arguably result in models that do not generalize well.

\setcode{utf8}

\begin{table*}
    \centering
    \scalebox{0.72}{%
     \begin{tabular}{l|lr}
     \toprule
     \multicolumn{1}{c}{\small\textbf{Type}} & \multicolumn{1}{c}{\small\textbf{Translation}} & \multicolumn{1}{c}{\small\textbf{Arabic}} \\ \hline
          \small \textbf{Reference} & \small \textbf{Wall Street records largest losses in 6 weeks}  & \small \<وول ستريت تسجل أكبر خسائر في 6 أسابيع> \\
          \small Paraphrase & \small \colorbox{green}{Losses in Wall Street are the highest} in 6 weeks & \small \<خسائر في وول ستريت هي الأعلى في ستة اسابيع	> \\
          \small Contradiction & \small \colorbox{pink}{Profits} in Wall Street in the last six weeks  & \small \<مكاسب في وول ستريت في الاسابيع الستة الاخيرة> \\
          \hline
          \small \textbf{Reference} & \small \textbf{Death of a journalist who reported on Russian Mercenaries}  & \small \<وفاة صحفي كتب عن المرتزقة الروس في سوريا في ظروف غامضة	> \\
          & \small \textbf{in Syria in mysterious circumstances} \\
          \small Paraphrase & \small Death of a journalist \colorbox{green}{in mysterious circumstances after he reported} & \small \<وفاة صحفى فى ظروف غامضة بعد أن كتب عن المرتزقة الروس فى سوريا> \\
          & \small \colorbox{green}{on} Russian Mercenaries in Syria \\
          \small Contradiction & \small Death of a journalist \colorbox{pink}{after battling with illness}  & \small \<وفاة صحفى بعد صراع مع المرض	> \\
          \hline
          \small \textbf{Reference} & \small \textbf{5.5 Billion withdrawn from emerging markets by investors in one week}  & \small \<5.5 مليار دولار سحوبات المستثمرين من الأسواق الناشئة بأسبوع> \\
          \small Paraphrase & \small \colorbox{green}{Nearly 6 Billion} withdrawn in a week from emerging markets  & \small \<قرابة ستة مليار دولار سحوبات أسبوع في الأسواق الناشئة> \\
          \small Contradiction & \small  \colorbox{pink}{Almost a million} in withdrawals from emerging markets  & \small \<سحوبات حوالي المليون في الأسواق الناشئة> \\
     \end{tabular}%
     }
     \caption{\small{Examples of modifications by annotators.  Green highlights a change in line with reference. Red highlights a conflicting part of the sentence with the reference sentence.}}
     \label{tab:examples}
 \end{table*}

\section{The corpus}
In this part, we describe our Arabic News Stance (ANS) corpus.\footnote{Data available at: \href{https://github.com/latynt/ans}{https://github.com/latynt/ans}} We derived two perspectives of the corpus suitable for claim verification and stance classification. Please refer to Appendix~\ref{data_statement} to read our data statement about the corpus.
\label{sec:data}

\subsection{Data Collection}
\label{sec:data_collection}
In contrast to \citet{Baly2018-ub} and more in line with \citet{Thorne2018-iw}, we start with true news titles (reference) and generate fake/true claims from them. The corpus generating process can be summarized in two stages: 1) generating true/false modifications of existing news titles through crowd-sourcing; and 2) validating the generated claims by annotating them in a separate phase.   

We derive our corpus by sampling a subset of news titles from the most recent version of the Arabic News Texts (ANT) corpus \citep{Chouigui2017-bu}; A collection of Arabic news from multiple news media sources in the Middle East. The dataset was suitable for our task as it covers several topics of news (politics, sports, \emph{etc.}) sourced from several credible mainstream news outlets (BBC, CNN, Al Arabiya, \emph{etc.}).  The following is an example of a news title from this dataset:

\begin{center}
\begin{tabular}{c}
\small

\<
حقائق سقوط صخرة تزن 100 كغ من الحائط
>
\\
\<
 الغربي بالقدس
>
\\
\emph{\space ``Facts about the falling of a boulder} \\
\emph{weighing 100 kg. of the west wall in Jerusalem.''} \\
\\
\end{tabular}
\end{center}

\paragraph{Generating true/false claims}

We used crowd-sourcing to generate true/false claims.  Starting from a news title, we recruited annotators to modify each news title into a new claim.  For true claims, annotators were asked to paraphrase the original sentence by changing its syntax and wording while maintaining the integrity of the information.  We allowed for the use of world knowledge to modify the information.  For example, replacing cities with countries and celebrities and politicians with their nationalities.  

For false claims, to insure that the modification results in meaningful mutation of the semantic information, the instructions (Table~\ref{tab:guides}) stated that the modified sentence should contradict the original title in such a way that both cannot simultaneously be true in the same context. Annotators were asked to avoid simple negation and were encouraged to use different strategies for modifying the sentences. Our analysis of a sample of the collected data showed that different annotators utilized different strategies at different rates. For example, some annotators predominantly altered years, counts and locations that appeared in the original titles while others modified the semantics of the modified sentences to have opposite meaning (detained vs. released, supported vs. opposed, etc.).

\paragraph{}
We relied on Amazon Mechanical Turk\footnote{https://www.mturk.com} and Upwork \footnote{https://www.upwork.com/} to recruit annotators.  We only considered Arabic native speakers for news title rewriting.  All annotators had to pass a language qualification test similar to our task. Data was randomly assigned to annotators in batches of 500. To insure the quality of the generated data, we sampled data during the annotation from each batch and re-annotated any batch containing errors in more than 10\% of the sample by resending the batch to the annotator after explaining the errors.  See Table~\ref{tab:examples} for examples of generated claims using different modification strategies.

\subsection{Data Validation And Analysis}

To evaluate the quality of our data, we performed a second round of annotation on the generated news titles.  We derived a new task in which annotators were presented with a pair of sentences and asked to supply a hypothesis about how they are semantically related.  This task is related to the the semantic concepts of entailment and contradiction \citep{Katz1972-iu, Bowman2015-ks} but with the aim of validating our generated ture/fake claims. We highlight a notable difference compared to other work on stance classification. In contrast to the commonly used four classes adopted in other datasets \footnote{For example: Fake News Challenge (FNC)} \textit{(agree, contradict, discuss, unrelated)}, we elect to merge labels \textit{(discuss, unrelated)} into one \textit{(other/not enough information)} resulting in three classes -- \textit{paraphrase, contradiction, other/not enough information} for each pair of news titles.  Our motivation is that despite the general value of discriminating between irrelevant documents\footnote{Documents in this case refer to sentences but could be any body of text. Hence, in this work we use both terms interchangeably.} (unrelated) and documents that are related to the claim but do not make a stance about the claim (discuss), both classes represent the same position in the context of stance prediction.  We, therefore, treat them as one class. We found that this is also similar to the approach by \citet{Thorne2018-iw}. 

\begin{table}
    \centering
    \small
    \begin{tabular}{lrr}
    \toprule
    \textbf{Number of Annotators} & \textbf{\# } & \textbf{\%} \\
        \phantom{ab}3 & 2594 & 60.9\%\\ 
        \phantom{ab}4 & 1239 & 29.1\%\\
        \phantom{ab}5 & 426 & 10.0\%\\
    \hline
    \textbf{Annotator Labels Overlap} & \textbf{\# } & \textbf{\%}  \\
         \phantom{ab}\textless\space75\% &  470 & 11.0\%\\
         \phantom{ab}75 - 99\% & 210 & 4.9\% \\
         \phantom{ab}100\% & 3579 & 84.0\% \\ 
    \hline
    \textbf{Majority/Author Labels Overlap} & \textbf{\# } & \textbf{\%} \\
        \phantom{ab}Majority Label = Author's Label  & 3766 & 99.4\%\\ 
        \phantom{ab}Majority Label $ \neq $ Author's Label  & 23 & 0.6\%\\
    \hline
    \textbf{Fleiss} $k$ &  &  \\
        \phantom{ab}3 total annotators  &  & 0.83\\ 
        \phantom{ab}4 total annotators  &  & 0.81\\ 
        \phantom{ab}5 total annotators  &  & 0.83\\ 
    \bottomrule
    \end{tabular}

    \caption{Statistics for the annotation results.  The author's label is the label obtained from the worker who rewrote the news title. Majority label is the consensus of 75\% or higher of the annotators.}
    \label{tab:perc_agreement}
\end{table}

\paragraph{}
To present annotators with a small set of the third class (other), we first considered randomly pairing news titles from our corpus. We hypothesized that randomly paired news titles will be discussing unrelated news and would naturally be assigned the label \textit{other} by annotators.  However and upon examining examples of this method, we noticed that telling the \textit(other) class apart from the two classes was dis-proportionally trivial since the randomly paired sentences differed significantly (discussed different topics and contained no overlapping words) compared to pairs from the \textit{paraphrase, contradict} classes.  Predicting this class, therefore, can be reduced to checking for the absence of overlap in words from the paired titles. As an alternative selection criteria to random pairing, we used a similarity metric to select pairs that look more similar.  We calculate the F1 score of overlapping ngrams in the paired titles weighted by the ngram size similar to \citet{Trinh2018-ti}.  In our case however, we consider ngrams at the character level given the short length of the sentences. We included ngrams of size 2 to 6 and set the minimum score to 0.1.

A total of 4,259 pairs were labeled by 3 to 5 annotators. We considered the author's rewritten sentences as labels (for the \textit{paraphrase} and \textit{contradict} classes). Table~\ref{tab:perc_agreement} shows the annotation statistics. The Fleiss $k$ scores (calculated separately for examples labeled by 3, 4 and 5 annotators) show overall a very high level of agreement $(>0.81)$ suggesting that the quality of the dataset is sufficiently high.  For the final data, we included only pairs with inter annotator agreement of 75\% or higher, hence, dismissing all data with 2 out of 3 majority vote or worse. 

\begin{figure}[h]
    \centering 
    \includegraphics[width=0.45\textwidth]{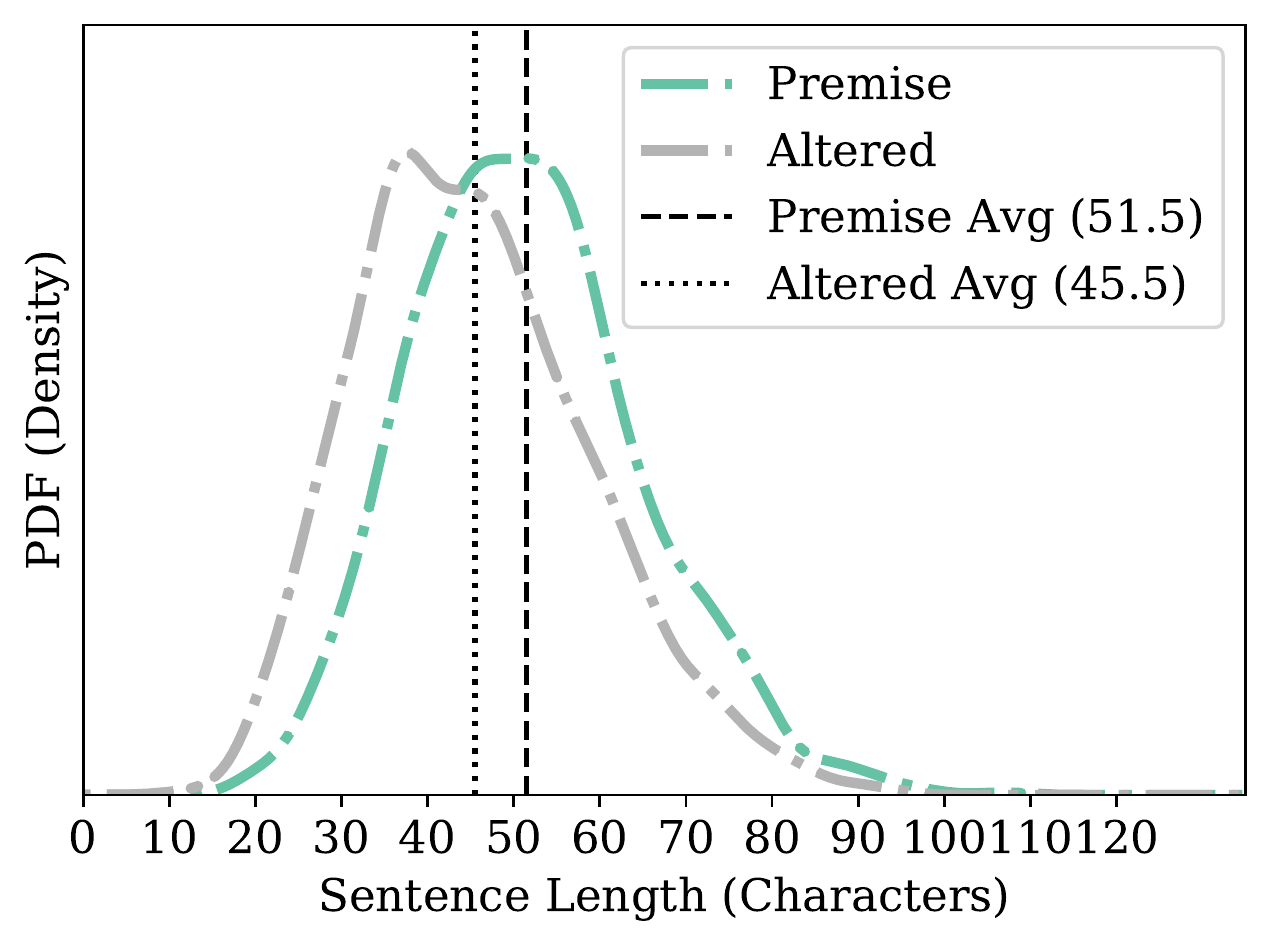}
    \caption{Length of sentences in dataset (rewritten vs. reference)}
    \label{fig:length_kdes}
\end{figure}

\begin{figure}[h]
    \centering 
    \includegraphics[width=0.45\textwidth]{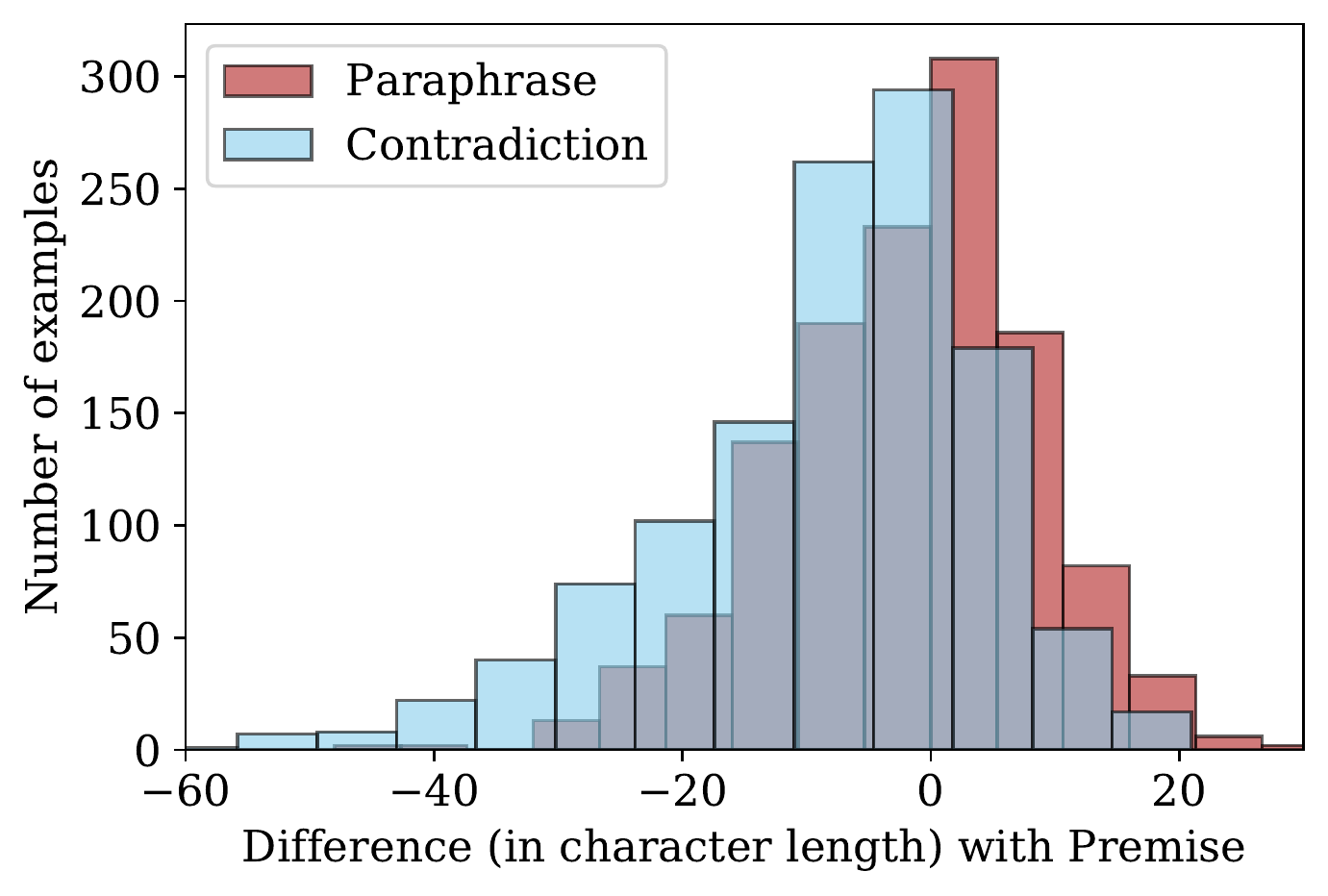}
    \caption{Comparison in rewritten sentences}
    \label{fig:length_diff}
\end{figure}

Figures~\ref{fig:length_kdes} and~\ref{fig:length_diff} provide some details about the length of the written claims in the final dataset compared to the original reference sentences. We noticed that on average, claims are shorter than the original references with contradicting claims being shorter than paraphrasing claims. This could be due to workers aiming to minimize time spent per each example. Another likely explanation is the fact that contradicting a statements by replacing or removing key words is easier than paraphrasing a statement.

\section{Experiments}
In this section, to demonstrate the utility of the corpus, we derive two tasks useful for evaluating news veracity and stance prediction and develop two baselines to evaluate on the proposed tasks.  We describe the proposed tasks and details of the baselines in this section and the results in section~\ref{sec:res}. 

\subsection{Tasks}

\paragraph{Claim Only Verification:}
\label{subsec:task_claim}
In this setting, we explore the task of verifying claims based only on information in the claims themselves. In our corpus, we assess the veracity of a claim $c_i$ from our corpus $D$ based solely on the textual information of the claim. The task is, hence, a binary classification where an estimator needs to map an input to a label $Y$ which can be either \emph{fake} or \emph{not fake}:

\begin{align*}
    p(Y|c_i) , \qquad c_i \in D
\end{align*}

\noindent We consider all original news titles (reference sentences) in our corpus to belong to the \emph{not fake} class. We rely on the fact that the reference sentences originated from reputable mainstream media in the Middle East.  Our \emph{fake} class examples consist of the sentences corrupted by annotators that passed the data validation process described in Section~\ref{sec:data_collection}.  Table~\ref{tab:Claim Verification} shows the distribution of classes for this task.

\begin{table}
    \centering
    \begin{tabular}{lrr}
    \toprule
    % & \multicolumn{2}{c}{Data}\\
    \textbf{Class} & \textbf{\# } & \textbf{\%} \\ \hline
        Not Fake  & 3072 & 67.6\%\\ 
        Fake  & 1475 & 32.4\%\\
    \bottomrule
    \end{tabular}
    \caption{ \small  Class distribution for claim verification. (\#: total number of examples. \%: percentage of all data)}
    \label{tab:Claim Verification}
\end{table}

 It is important to discern the limited scope in defining news veracity in this work: the incorrectness of the corrupted sentence is not a universal statement about the claim.  We note the fact that several of the corrupted sentences can be factual/not fake in other contexts.  As such, we consider them fake in regards to the related event/context - in this case our reference sentence (original news titles). Further analysis exposed two instances where the modified sentences matched other original news titles. Both examples were excluded from the corpus for this task. However, such cases hint at the limits of claim verification using claim text only.  We further explore this in section~\ref{sec:res} and share some insights.

\paragraph{Stance Prediction}

This task is a direct reflection of our annotation process.  Given a reference sentence $r_i$ and a claim $c_i$, predict the label $Y$ (Agree, Contradict, Other/Not enough information) from the claim/reference pair $(c_i, r_i)$. 

\begin{align*}
    p(Y|c_i, r_i) , \qquad (c_i, r_i) \in D
\end{align*}

 Table~\ref{tab:Stance Prediction} shows the distribution of classes in our corpus for the stance prediction task.

\begin{table}
    \centering
    \begin{tabular}{lrr}
    \toprule
    % & \multicolumn{2}{c}{Data}\\
    \textbf{Class} & \textbf{\# } & \textbf{\%} \\ \hline
        Disagree  & 2399 & 63.4\%\\ 
        Agree  & 1301 & 34.4\%\\
        Other & 86 & 2.3\%\\
    \bottomrule
    \end{tabular}
    \caption{\small Class distribution for stance prediction. (\#: total number of examples. \%: percentage of all data)}
    \label{tab:Stance Prediction}
\end{table}

\subsection{Methods}

We evaluate two baselines on both tasks. For modeling, we considered two classes of models that have been largely adopted by the NLP community.  The models are described in the next section.

\begin{table*}[h!]
\centering
\small
\begin{tabular}{ l@{\hskip .5in}cccc@{\hskip .5in}cccc@{\hskip .2in}cccc }
\toprule 
\multicolumn{1}{c}{} & \multicolumn{4}{c}{(dev)} & \multicolumn{4}{c}{(test)} \\ \hline
\textbf{Claim Verification} & 
\textbf{Acc.} & \textbf{Prec.} & \textbf{Rec.} & \textbf{$F_1$} &
\textbf{Acc.} & \textbf{Prec.} & \textbf{Rec.} & \textbf{$F_1$} \\ \hline

\\[-.5em] %add additional space from above
Majority Class & 68.1&34.1&50.0&40.5&67.1&33.6&50.0&40.2 \\ \hdashline
\\[-.5em] %add additional space from above
LSTM character level\\
\phantom{888} \textit{char,   10(emb),  100(hid),    0(dropout)} & 70.2&65.7&56.8&55.4&67.3&60.2&54.6&52.5 \\
\phantom{888} \textit{char,   10(emb),  100(hid), 30.0(dropout)} & 70.6&67.9&56.5&54.6&67.8&61.3&55.1&53.1 \\
\hdashline
LSTM word level\\
\phantom{888} \textit{word,   50(emb),   50(hid),    0(dropout)} & 68.1&60.4&54.8&52.9&65.8&57.2&53.9&52.4 \\
\phantom{888} \textit{word,   50(emb),  100(hid),    0(dropout)} & 68.6&61.8&56.4&55.5&64.5&55.4&53.3&52.1 \\
%\hdashline

\hline
\textbf{Stance Prediction} \\ \hline

%data
\\[-.5em] %add additional space from above
Majority Class & 62.4&20.8&33.3&25.6&63.8&21.3&33.3&26.0 \\ \hdashline
\\[-.5em] %add additional space from above

LSTM character level \\
\phantom{888} \textit{char,   10(emb),   50(hid),    0(dropout)} & 62.2&20.7&33.3&25.6&64.4&21.5&33.3&26.1 \\
\phantom{888} \textit{char,   50(emb),   50(hid),    0(dropout)} & 62.4&20.8&33.3&25.6&64.1&21.4&33.3&26.0 \\
\phantom{888} \textit{char,   50(emb),   50(hid), 30.0(dropout)} & 62.5&43.0&33.7&26.6&64.4&46.4&34.0&27.5 \\
\hdashline
LSTM word level \\
\phantom{888} \textit{word,   10(emb),   50(hid),    0(dropout)} & 62.1&38.7&39.2&38.8&62.0&37.8&38.1&37.8 \\
\phantom{888} \textit{word,   50(emb),   50(hid), 30.0(dropout)} & 63.0&39.9&40.7&40.3&59.8&37.4&38.2&37.8 \\

\bottomrule
\end{tabular}
\caption{\label{res-task1} Results for the claim verification and stance prediction Tasks. }
\end{table*}

\paragraph{Recurrent Perspective Matching:}
Our first baseline is a simple RNN model that uses Long Short Term Memory (LSTM) \citep{Hochreiter1997-fh} as the main building block to encode the input.  LSTM models encode the input sequentially and can model temporal dependencies useful to semantic tasks.  In our implementation for both tasks, we consider both character level and word level representations of the input sentence(s) separately. In each case, we represent every input word/character with a unique $d$-dimensional vector that is learned during training. These vectors are then passed through the LSTM layer in sequence and the output of the last step (at the end of the sentence) is used as the encoding of the sentence(s). For the claim verification task, the claim encoding $\overrightarrow{h}_t$ can be described by:

\begin{align*}
\overrightarrow{h}_t &= \overrightarrow{LSTM}(\overrightarrow{h}_{t-1}, x_t) \qquad t = 1,...,M_i
\end{align*}

\noindent Where $M_i$ is the length size of the sentence corresponding to example $i$ and $x_t$ is the character/word at position $t$. 

\noindent In stance prediction, the input consists of a pair of sentences (reference $r$, claim $c$). Each is encoded using the same LSTM layer to obtain their encoding:

\begin{align*}
\overrightarrow{r}_t &= \overrightarrow{LSTM}(\overrightarrow{r}_{t-1}, x_t) \qquad t = 1,...,M_i^r \\
\overrightarrow{c}_t &= \overrightarrow{LSTM}(\overrightarrow{c}_{t-1}, x_t) \qquad t = 1,...,M_i^k
\end{align*}

\noindent To obtain the interaction representation $\overrightarrow{h}_t$, $\overrightarrow{\text{r}_{t}}$ and $\overrightarrow{\text{c}_{t}}$ are multiplied element-wise.  We experimented with \emph{cosine} similarity and concatenation and found the element-wise multiplication and concatenation to work slightly better than \emph{cosine} similarity: 

\begin{align*}
\overrightarrow{h}_t &= (\overrightarrow{\text{r}_{t}} \circ \overrightarrow{k}_{t})
\end{align*}

\noindent The resulting encoding in both tasks $\overrightarrow{h}_t$ is then passed through a linear layer with non-linearity (\emph{ReLu}) followed by a $softmax$ function to convert the output to probabilities for each class:

\begin{align*}
\small
    & p(Y=c|h_i) = softmax(ReLu(W_c\overrightarrow{h_i}+b_c))
\end{align*}

\noindent $W_c$ and $b_c$ are learnable parameters associated with each class $c$ in the corresponding task.  

\noindent Prediction in both tasks is done by selecting the label with the highest probability: 

\begin{align*}
\small
    & \argmax_{c} p(Y=c|h_i)
\end{align*}

\paragraph{Pretrained Transformer:}
\label{sec:pretraining}

Pretraining and transfer learning \citep{Devlin2018-ka, Peters2018-zq, Radford2019-ob} has recently gained attention as a popular approach to acquiring universal linguistic features useful in many downstream NLP tasks and was shown to be successful in improving on the state of the art in many downstream NLP tasks with minimal fine-tuning. \citet{Lv2019-sp} have successfully explored BERT for the task of fake news detection in English and proposed an extension that improves on fine-tuned BERT.  In addition to the aforementioned supervised methods, we evaluate BERT \citep{Devlin2018-ka} on both tasks in our corpus. We are not aware of any other work that explored pretraining for claim verification and stance prediction in Arabic.  

BERT is based on the Transformer model first introduced by \citet{Vaswani2017-vp}. Transformer-based models have recently become common in many NLP tasks including question answering and entailment classification \citep{DBLP:journals/corr/abs-1810-04805, Radford2018ImprovingLU}. For both tasks, we utilize a publicly available implementation that has been trained on a multilingual dataset including Arabic.\footnote{We use BERT-Base, Multilingual Cased: 104 languages, 12-layer, 768-hidden, 12-heads, 110M parameters}.  We elect to adhere to the proposed approach recommended by \citet{Devlin2018-ka} for future reproducibility.   Since our implementation is identical to the one provided by the authors, we will omit the detailed description of the model architecture and refer readers to \citep{Vaswani2017-vp}\footnote{See also:\newline http://nlp.seas.harvard.edu/2018/04/03/attention.html}.

\begin{table}[h]
\centering
\small
%\scalebox{0.85}{%
\renewcommand{\arraystretch}{1.2}
\begin{tabular}{l@{\hskip .5in}ccc}
\hline
\textbf{Task} &  \textbf{Prec.} & \textbf{Rec.} & \textbf{$F_1$} \\
\hline
%No & 63.9 / 21.3 / 33.3 / 25.9 & 67.1 / 33.6 / 50.0 / 40.2 \\
\textbf{Claim Verification} & \\
\phantom{888} \textit{Fake} & 51 & 55 & 53 \\
\phantom{888} \textit{Not Fake} & 77 & 75 & 76 \\
\phantom{888} \textit{\textbf{Macro Avg.}} & \textbf{64.1} & \textbf{64.6} & \textbf{64.3} \\
\hdashline
\textbf{Stance Detection} & \\
\phantom{888} \textit{Agree} & 65 & 63 & 64 \\
\phantom{888} \textit{Disagree} & 80 & 81 & 80\\
\phantom{888} \textit{Other} & 86 & 86 & 86\\
\phantom{888} \textit{\textbf{Macro Avg.}} &  \textbf{76.8} & \textbf{76.6} & \textbf{76.7} \\
\hline
\end{tabular}%
%}
\caption{\label{res-pretraining} Results of using pretraining (BERT) on claim verification and stance prediction tasks.}
\end{table}

\begin{table*}[h]
    \centering
    \scalebox{0.8}{%
     \begin{tabular}{l:llr}
     \toprule
     %\multicolumn{2}{c}{\textbf{Example of the data}} \\    \hline
     \multicolumn{1}{c}{\small\textbf{Prediction}} & \multicolumn{1}{c}{\small\textbf{Label}} & \multicolumn{1}{c}{\small\textbf{Translation}} & \multicolumn{1}{c}{\small\textbf{Arabic}} \\ \hline
          \small \textbf{ Fake} &  \small  \textbf{Fake} & \small Historic agreement between Europe and Japan to \colorbox{yellow}{support} trump  & \small \<اتفاق تاريخي بين أوروبا و اليابان لمساعدة ترامب> \\ 
          \small  \textbf{\colorbox{pink}{Fake}} & \small \textbf{True} &   \small Historic agreement between Europe and Japan to \colorbox{yellow}{confront} trump  & \small \<اتفاق تاريخي بين أوروبا و اليابان لمواجهة ترامب> \\
          \hdashline
          \small  \textbf{\colorbox{pink}{True}} & \small \textbf{Fake} &   \small First women's interest channel in Gaza \colorbox{yellow}{soon to see the light}  & \small \<أول قناة تلفزيونية نسائية في غزة تظهر للنور قريبا>  \\
          \small \textbf{ True} &  \small  \textbf{True} & \small First women's interest channel in Gaza \colorbox{yellow}{faces uncertain fate}  & \small \<أول قناة تلفزيونية نسائية في غزة تواجه مصيرا مجهولا>  \\
          \hdashline
          \small  \textbf{\colorbox{pink}{Fake}} & \small \textbf{True} &   \small Ethiopia \colorbox{yellow}{assures} Egypt of its Nile share & \small \<أثيوبيا تؤكد حرصها على حصة مصر بالنيل> \\
          \small \textbf{ Fake} &  \small  \textbf{Fake} & \small Ethiopia \colorbox{yellow}{apathetic} about Egypt's right of the Nile water & \small \<أثيوبيا غير معنية بحصة مصر من مياه النيل>  \\
     \end{tabular}%
     }
     \caption{\small{Examples of claim verification task predictions using fine-tuned BERT highlighting the model's invariant labels for similar sentences with different meanings.}}
     \label{tab:results_claim_samples}
 \end{table*}
 
\section{Results}
\label{sec:res}

For the recurrent perspective models, we trained all models for 100 epochs using Adam optimizer \cite{Kingma2014-su} with 0.001 learning rate. We conducted hyper-parameter tuning on the development set. For the pretrained BERT model, we fine-tune on our data for 3 epochs using BERT BPE units.

\paragraph{} Table~\ref{res-task1} shows the top results of all experiments for both tasks. We report the accuracy and $F_1$ (Macro unweighted average). In the claim verification task, results show that in general, word based models perform comparably to character based models but we note that all results do not provide significant gains (53.1 vs. 40.2 $F_1$) compared to the baseline (majority class) which could be explained by the small training data size but might hint at an ill-defined task.  We explore this further below.  In the stance prediction task, experiments show word based models outperform character based models (37.8 vs 27.5 $F_1$). This could be due to the limited size of our corpus which is not sufficient for character based models to learn words and phrases from scratch and capture the semantic representation needed for stance prediction.

Results for the pretraining experiments (shown in Table~\ref{res-pretraining}) show significant improvement of the pretrained model over the models trained only on our corpus.  This is similar to findings by \citet{Lv2019-sp}. However, the improvement is dis-proportionally larger in the stance prediction task (76.7 vs. 37.8 $F_1$) and the large gains do not carry over to the claim verification task (64.3 vs. 53.1 $F_1$).  The imbalance in gains also confirms our intuition about the limitation of claim only verification which we discuss next.

\paragraph{Limits Of Claim Only Verification:}
We briefly mentioned in Section~\ref{subsec:task_claim} the limited scope of claim verification in a setting were the decision about the veracity of a claim can be made using only the text of the claim.  We hypothesize that the task might not be learnable through a direct mapping from the claim text to the veracity space. Given that the initial results of the fine-tuned BERT model supported this intuition, we elected to manually inspect a sample of the predictions and noticed that in many cases the model was predicting the same label for claims that look similar but are semantically different. We share a sample of these cases in Table~\ref{tab:results_claim_samples}. This suggests that while the linguistic features learned during pretraining were useful for textual entailment (stance prediction task), the veracity of a claim cannot be made using only implicit world knowledge learned during pretraining. A simple example highlighting this limitation is the reference news title \<الذهب يصعد مع تراجع الدولار> \emph{``Gold prices increase amidst a falling dollar.'''} and its contradicted rewritten version "\<أسعار الذهب تهبط عالميا"> \emph{``Gold prices fall globally''}. Here, it is easy to argue that the contradiction can be true in another context and hence, a decision about the veracity of this claim should only be made in reference to a particular context/event. We believe that explicitly associating each claim with evidence or context is the more appropriate approach for claim verification.

\noindent These initial experiments suggest that discriminate models trained using claim only information might rely on biases in the topics, linguistic styles, tones and implicit world-knowledge learned from training data to make predictions. Results of the performance of such models could, therefore, be inflated if the training data is not uniformly distributed across languages, topics, writing styles, political ideologies etc. While we believe that our dataset collection process which yields classes that share the same distribution of topics and news sources mitigated these types of biases, we also note that the annotation process and human factor introduced other types of biases that could be present in the data.

\section{Conclusion}
In this work we presented a new publicly available corpus for textual entailment and its use in studying misinformation in the Arabic language.  We shared some insights about the creation of the corpus and the baselines developed to evaluate the corpus.  We further explored the use of pretraining \citep{Devlin2018-ka} and developed a strong baseline for our tasks.  Our experiments additionally shed light on the limits of "claim-only" misinformation detection methods that rely solely on the stated claims without use of accompanying evidence. We hope to explore this further in future work. As we plan to also explore the use of generated data in studying the robustness of misinformation detection methods against adversarial data with varying linguistic styles, political ideologies and world-knowledge. 

\section*{Acknowledgements}
We are grateful for Ayah Zirikly, Bart Desmet and René F. Kizilcec for their valuable insights, critical commentary and helpful discussions during the course of this work. We also thank our anonymous reviewers for their thoughtful comments and suggestions.

\bibliography{ans}
\bibliographystyle{acl_natbib}

\appendix

\section{Data Statement}
\label{data_statement}

In line with recent efforts addressing ethical issues that can result from the use of data and technology and following the recommendations of \citet{TACL1464}, we are sharing the following information that we believed is relevant to our dataset and the collection process.  We encourage future use of the data to include a summary of this information.

\subsection{Language Variety}

To study and build tools in the areas of stance prediction and claim verification. Data was selected from news titles and rewritten by annotators for the purpose of generating statements and statement pairs. Part of the dataset was a random subset of the ANT corpus which was created through web-crawling news sources in the Middle East.  As different tools and annotation were included in the creation of the data, we expect the distribution of topics, opinions and language to incorporate different types and levels of bias. To the best of our knowledge, the data is in Standard Arabic ('arb') with few exceptions such as abbreviations. At least Latin script ('Latn') is present.

\subsection{Annotator Demographic}

A total of 8 crowd-source workers mostly from the Middle East contributed to the annotations.  Annotators were selected based on their fluency in the Arabic language. Demographic information was not available at the time annotation for all recruited individuals. Of the information available, we are aware of at least 1 woman, 2 men and 3 individuals who are Arabic native speakers.

\subsection{Text Characteristics}

The dataset includes a subset of the news titles from ANT news corpus (v2.1) which included 5 news sources (BBC, Al Arabiya, CNN, Sky News, France24) and 6 categories (culture, economy, international news, Middle East, sport, technology) collected from February 2018 to October 2018.  Data also includes rewritten versions of the news titles by the annotators following the provided guidelines (see Table~\ref{tab:guides}).

\end{document}